%% file: main.tex
\documentclass[letterpaper, 10 pt, conference]{ieeeconf}  
\IEEEoverridecommandlockouts 
\usepackage{blindtext, graphicx}

\usepackage{caption}

\usepackage{xcolor}
\definecolor{cornflower_blue}{rgb}{0.392157, 0.584314, 0.929412}
\definecolor{teal}{rgb}{0, 0.6501961, 0.6501961}
\definecolor{poppy}{rgb}{0.890196, 0.32549, 0.207843}
\definecolor{forest}{rgb}{0.136719, 0.554688, 0.136719}
\definecolor{red}{rgb}{1.0, 0.0, 0.0}
\definecolor{green}{rgb}{0.0, 1.0, 0.0}
\definecolor{blue}{rgb}{0.0, 0.0, 1.0}

\newcommand{\newtext}[1]{#1} 

\newcommand{\etal}{\textit{et al}.}

\newcommand{\eg}{\textit{e}.\textit{g}.}

\usepackage{algorithm}
\usepackage{algpseudocode}
\usepackage{multirow}
\begin{document}

\title{Learning autonomous driving from aerial imagery}

\author{Varun Murali$^{1}$, Guy, Rosman$^{2}$, Sertac, Karaman$^{1}$ and Daniela Rus$^{3}$
\thanks{
*This work is supported by Toyota Research Institute (TRI).  It, however, reflects solely the opinions and conclusions of its authors and not TRI or any other Toyota entity.}
\thanks{$^{1}$MIT LIDS, $^{2}$TRI, $^{3}$MIT CSAIL.}
}
\maketitle

\begin{abstract}
In this work, we consider the problem of learning end to end perception to control for ground vehicles solely from aerial imagery. 
Photogrammetric simulators allow the synthesis of novel views through the transformation of pre-generated assets into novel views.
However, they have a large setup cost, require careful collection of data and often human effort to create usable simulators.
We use a Neural Radiance Field (NeRF) as an intermediate representation to synthesize novel views from the point of view of a ground vehicle.
These novel viewpoints can then be used for several downstream autonomous navigation applications.
In this work, we demonstrate the utility of novel view synthesis though the application of training a policy for end to end learning from images and depth data.
In a traditional real to sim to real framework, the collected data would be transformed into a visual simulator which could then be used to generate novel views.
In contrast, using a NeRF allows a compact representation and the ability to optimize over the parameters of the visual simulator as more data is gathered in the environment.
We demonstrate the efficacy of our method in a custom built mini-city environment through the deployment of imitation policies on robotic cars.
We additionally consider the task of place localization and demonstrate that our method is able to relocalize the car in the real world.
\end{abstract}

\IEEEpeerreviewmaketitle

\input{introduction}

\input{related-work}
\input{problem}

\input{experiments}

\section{Conclusion}

In summary, we have presented an innovative approach for teaching ground robots visuo-motor policies and localization, using solely bird's eye view images and their corresponding poses as input. 
This methodology was successfully applied and tested on a 1/10th scale environment and car within the minicity platform. 
Our findings suggest the potential to enable the acquisition of autonomous driving capabilities with a limited quantity of aerial imagery, in contrast to the conventional approach that relies on extensive datasets. 
\newtext{Our results also show that our method can also closely approximate the results of using imagery from the same domain alleviating the necessity of data from the same view.}
In future work we aim to extend this methodology to full-scale vehicles, operating in large outdoor settings, by harnessing the power of open source satellite data.
We can also consider introducing other functions such as dynamic actors into the view synthesis using a time-dependent nerf.

\bibliographystyle{IEEEtran}  
\bibliography{bibliography.bib}
\end{document}

%% file: introduction.tex
\section{Introduction}
\label{sec:introduction}

\begin{figure}[!ht]
\centering
\includegraphics[width=0.5\textwidth]{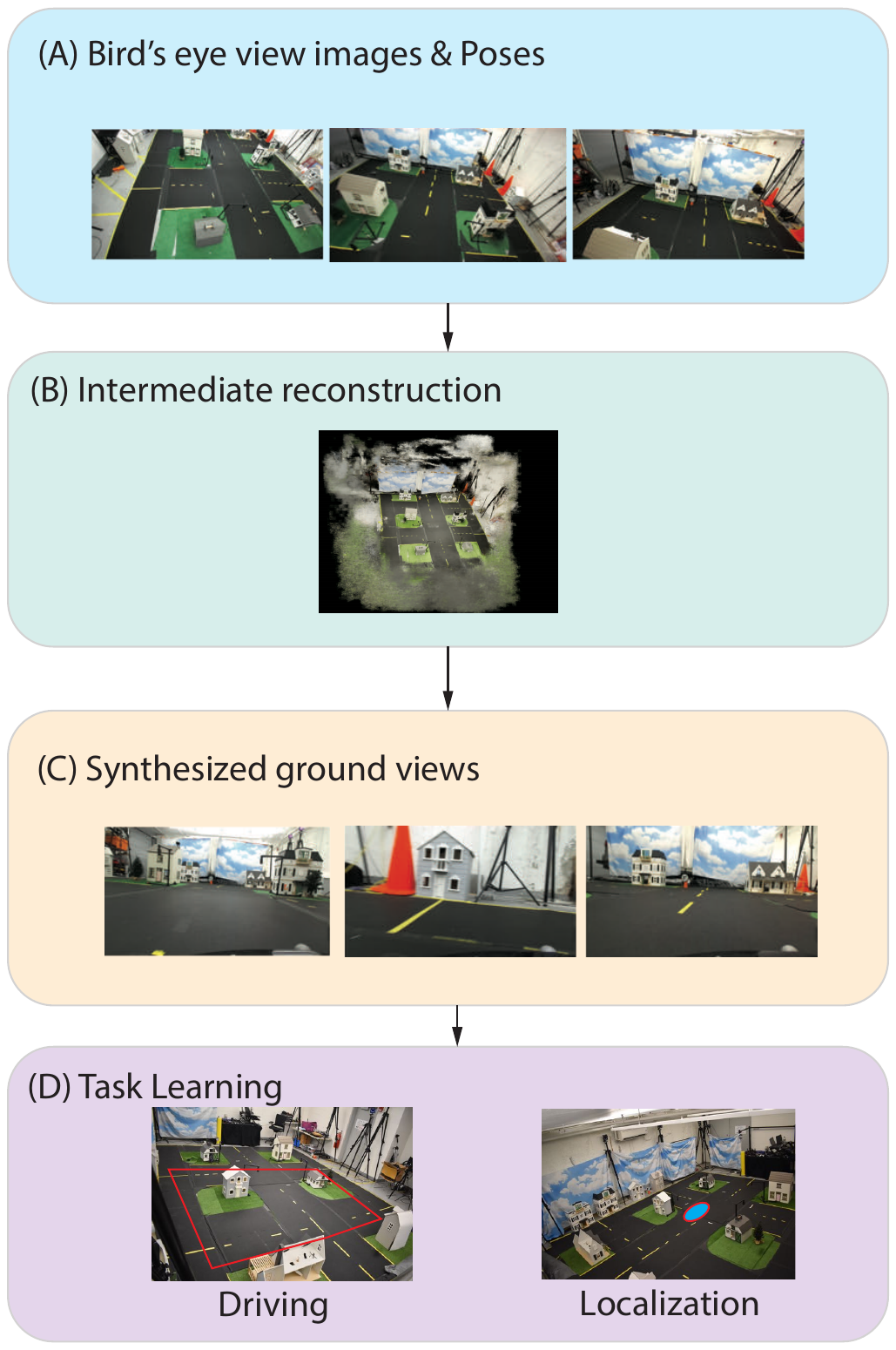}
\caption{An overview of the method presented in this work. First, we assume that we are given aerial images and their corresponding poses (A). We then reconstruct a photogrammetric model (B) of the world using BEV images and use this intermediate representation which can be used to query  ground robot views for desired poses (C). We use the synthesize images to learn policies that can be directly deployed on real vehicles (D).
}
\label{fig:method-overview}
\end{figure}

With the rising popularity of self-driving cars and the need for high definition maps for navigation, it is becoming ever necessary to gather data for training autonomous driving on urban roads.
This requires collecting continuous driving data and is limited to regions which are heavily trafficked or allowable by the rules of the road.
On the other hand, aerial imagery captures larger regions from a different viewpoint which can be transformed into the viewpoint of a ground robot.
This allows the capture of edge cases or visual information that might be otherwise occluded cumbersome to represent in visual reconstructions or simulators.
Visual simulators have also grown increasingly in capability with the advancement of real time ray tracing solutions such as Nvidia DLSS and the availability of commercial photogrammetric tools such as RealityCapture \cite{RealityCapture}, Matterport \cite{Ramakrishnan2021-pw}.

Aerial images in contrast are readily available in the form of satellite imagery and are less cumbersome to collect since they are able to map multiple streets simultaneously. 
In this work, we operate under the real to sim to real framework where we transform images into novel viewpoints for data generation through the intermediate representation. 
Since this representation is also in itself learnt, we are able to store the representation in a compact way and optimize over parameters such as lighting, and locations of objects where new data is available.
In contrast, a traditional simulator (built through photogrammetry such as FlightGoggles \cite{guerra2019flightgoggles}) would require the transfer of imagery into visual assets that can then be transformed into novel views but updating it would require the generation of new assets or creation of a new database for photogrammetry. 
Photogrammetric simulators are then able to output various visual modalities including the camera imagery, depth imagery and other representations such as semantic segmentation and surface normals.
These can then be used to augment the training data available to train autonomous agents (such as in VISTA \cite{amini2020learning}) or allow agents to learn policies through trial in simulation (such as in AI habitat \cite{savva2019habitat}).
Such simulations allow us to consider data that would not otherwise be available in traditional human collected data, \eg driving off-road or near accident scenarios.
\newtext{Generative simulation~\cite{singh2023worldgen} is already playing a role in robotics applications across multiple domains.
Simulation engines are increasingly used to combine photogrammetric assets and generated ones to train machine learning algorithms but this requires a lot of data and engineering effort.
It is also unclear how to capture the day to day changes in the environment with high fidelity in such a simulation.
}

Learning representations of the physical world also alleviates the need to reason about occlusions during the synthesis of (depth and color) images  from the reconstruction.
In a traditional setting, the simulator would hold a object mesh of the physical environment and then project rays and reason about occlusions from a certain viewpoint to render the scene onto camera space. 
In contrast, a neural radiance field learns both the color along the ray and the density along the ray implicitly learning the occupancy map.
%
\newtext{For this work, we consider learning from these representations two problems in autonomous driving: (i) visual re-localization and (ii) end-to-end driving.
Visual (re-)localization is a challenging problem in infrastructure denied navigation.
Most driving applications require localization to downstream tasks.
End-to-end driving on the other hand attempts to implictly learn the localization.
While, these tasks are related in that localization could aid a motion planning algorithm to follow trajectories by considering obstacles in the environment, in this work the second task is learnt completely independently to demonstrate the possibility of implicitly reasoning about plans from only on-board imagery.
We stress here that the applications are cross-domain tasks i.e. taking BEV images as input and estimating pose or desired actions in a ground robot's view.
We also stress that while we use NeRF as the intermediate perceptual representation for cross-view tasking, the methods presented in this work are general enough to be extended to other methods such as Gaussian Splatting~\cite{kerbl20233d}.
Fig~\ref{fig:method-overview} outlines the proposed components of this work.
}

In summary, in this work we represent the transformation of a collection of images from a bird's eye view (BEV) into a ground view (rgb imagery) using a neural radiance field as an intermediate representation. 
We then use the generated ground views to train an imitation policy using a pure-pursuit controller to learn a mapping from the robot's onboard camera to a steering and velocity command.
We demonstrate the utility of our method through real world experiments on one-tenth scale cars in a miniature city environment by transferring the learned policies directly onto the vehicles.

%% file: related-work.tex
\section{Related Work}
Our work intersects with several topics of research in robotics and robotic perception.
In particular, we leverage ideas from multi-view stereo from aerial images and novel view synthesis from neural rendering fields.
We also leverage ideas from reasoning across varied viewpoints and photogrammetric real-to-sim-to-real pipelines which generate photorealistic imagery to train machine learning algorithms.
Finally, we demonstrate the efficacy of our approach through the tasks of (i) visuo-motor policy learning and (ii) visual place recoginition.
\subsection{Multi-View reconstruction from aerial images}

Several works have considered the problem of generating dense reconstruction from satellite and aerial images such as \cite{Cabezas2014,VisSat-2019, schoenberger2016sfm,beyer2018ames, bosch2016multiple, d2012dense, gong2019dsm}. 
These works focus on matching features across the images and aligning them to minimize a photometric error to recover a dense surface model. 
Another set of related works involved ground-and-aerial navigation and planning \cite{Chen2021-fm,Downes2022-wn}.
Since these works usually involve cumbersome feature detection, matching and global bundle adjustment to generate the model they are computationally expensive to compute and would require an additional meshing procedure to store the generated surface models. Additional reasoning is often needed to cull occluded points and vary parameters like lighting to produce realistic images that are actionable and artifact-free. Leveraging neural radiance fields as we do in this work help alleviate some of these challenges.

\subsection{Novel View Synthesis}
Neural radiance fields were first presented by Mildenhall et al. \cite{Mildenhall2021-lx} to learn the mapping of a projected ray to the colors and the log density along the ray. 
Since the initial appearance of the neural radiance fields, several works has considered improvements such as relighting the scene \cite{Toschi2023-qw}, relaxing the requirement of accurate poses \cite{Lin2021-ku}, extensions to style transfer and near-real time field generation \cite{muller2022instant}.
\cite{mari2022sat, muller2022instant}
Recent work also considers the idea of regularizing the predicted densities in order to improve the quality of the depth maps \cite{Wu2023-ry}.
Multi-view NeRFs have also been proposed to better utilize nearby reference views at the time of inference \cite{Chen2021-tr}.
\newtext{City scale reconstructions using NeRFs have also been considered \cite{tancik2022block}.
This effectively stores the problem of storing large scenes but is trained on similar views as the rendered views.
3D Gaussian splatting based approaches have also been used as an alternative to neural radiance fields to learn dense representations from multi-view camera imagery \cite{kerbl20233d}.
In this work, we use NeRF as a backbone representation to store the perceptual information but stress that the proposed method is easily extensible to other representations.
}

\subsection{Representations for Cross-view Embodied Reasoning}
The problem of learning to act in the world based on a different view of the scene is been the focus of several research threads.
In trajectory generation and forecasting, several works have looked at birds-eye-views as an input for acting. 
\cite{gilitschenski2020deep} have learned a latent map representation from BEV images and embedded local cues for acting into it.
Adamkiewicz \etal \cite{Adamkiewicz2022-re} have leveraged NeRFs to create an occupancy estimate and planned in the resulting volume.
For localizing agents, \cite{Lin2021-ku} have demonstrated how to correct camera poses along with reconstructing the NeRF, whereas Moreau \etal \cite{moreau2022lens} proposed how to leverage NeRFs to enhance localization.
\newtext{Semantic representations for cross-view localization have also been studied \cite{chiu2018vr, miller2022stronger} where the geometric composition of semantically important objects in the scene are used to register aerial views to ground views.}

\subsection{Photogrammetric simulators}
Simulators such as FlightGoggles \cite{guerra2019flightgoggles} aim to solve the sim to real to sim by first generating the photogrammetric assets through software and human effort and then feeding it to a game engine such as Unity3D.
In this scenario, the assets are ``baked" into the simulator and the approximate ray tracing methods relevant in the computer graphics community can be used to relight the scene and synthesize images.
However, while such simulators are capable of fast generation of imagery, the assets themselves require high level of detail to store and are often expensive in terms of memory.
Any changes to the physical environment would also require repeating the asset capture process (depending on the granularity of the change).
Simulators such as VISTA alleviate the photogrammetric asset creation by generating only local transformations from pre-existing datasets.
While this allows for a diversity in generating training data, it does not allow the introduction of imagery from a drastically different view point.
\newtext{Photogrammetric simulation also often takes a lot of time to generate and requires capturing a lot of imagery for good performance.}

\subsection{Visuo-motor policies}
Prior work also considers learning policies that ingest vision and directly regress the desired robot control.
\cite{Levine2015-mu} address the question of training perception and control jointly and demonstrate the ability to learn manipulation tasks directly from imagery.
\cite{amini2019variational} use imitation learning to learn end-to-end driving given on-board imagery from the car and noisy localization estimate. \cite{bojarski2016end, xu2017end, hawke2020urban} demonstrate learning end-to-end driving from a large amount of driving video data.
\newtext{\cite{byravan2023nerf2real} use NeRFs to generate simulation environments from scenes by collecting data from a phone and then learn locmotion policies in this simulation.
Although the general application and idea is similar, in this work we utilize images from different domains i.e. BEV to train the NeRF and ground views for the simulation.}

\subsection{Visual place recognition}
Work has also considered the problem of localization from single images given a visual library or map \cite{Lowry2016-sx}.  
For instance, \cite{kendall2015posenet} learn to regress the camera position and orientation given a sequence of images and poses from the same viewpoint and same region.
Newer work has also considered learning or learnt feature representations to hash maps and estimate the pose from images \cite{Zhang2021-nl}.
In contrast to prior work, we directly leverage aerial imagery to perform cross view tasks that are relevant to autonomous driving. 
Specifically, we demonstrate the ability to localize the ego-vehicle and perform visuo-motor control onboard ground vehicles with only imagery from an aerial vehicle.

%% file: problem.tex
\section{Method}
\label{sec:method}

\newtext{In this section, we first detail the problem setup.
We then explain the procedure for generating the intermediate NeRF based simulation.
Then, we present two two learning problems that are relevant for autonomous driving.}

\begin{figure*}[!tbh]
\centering
\includegraphics[width=0.95\textwidth,trim=0cm 5cm 0cm 6cm,clip]{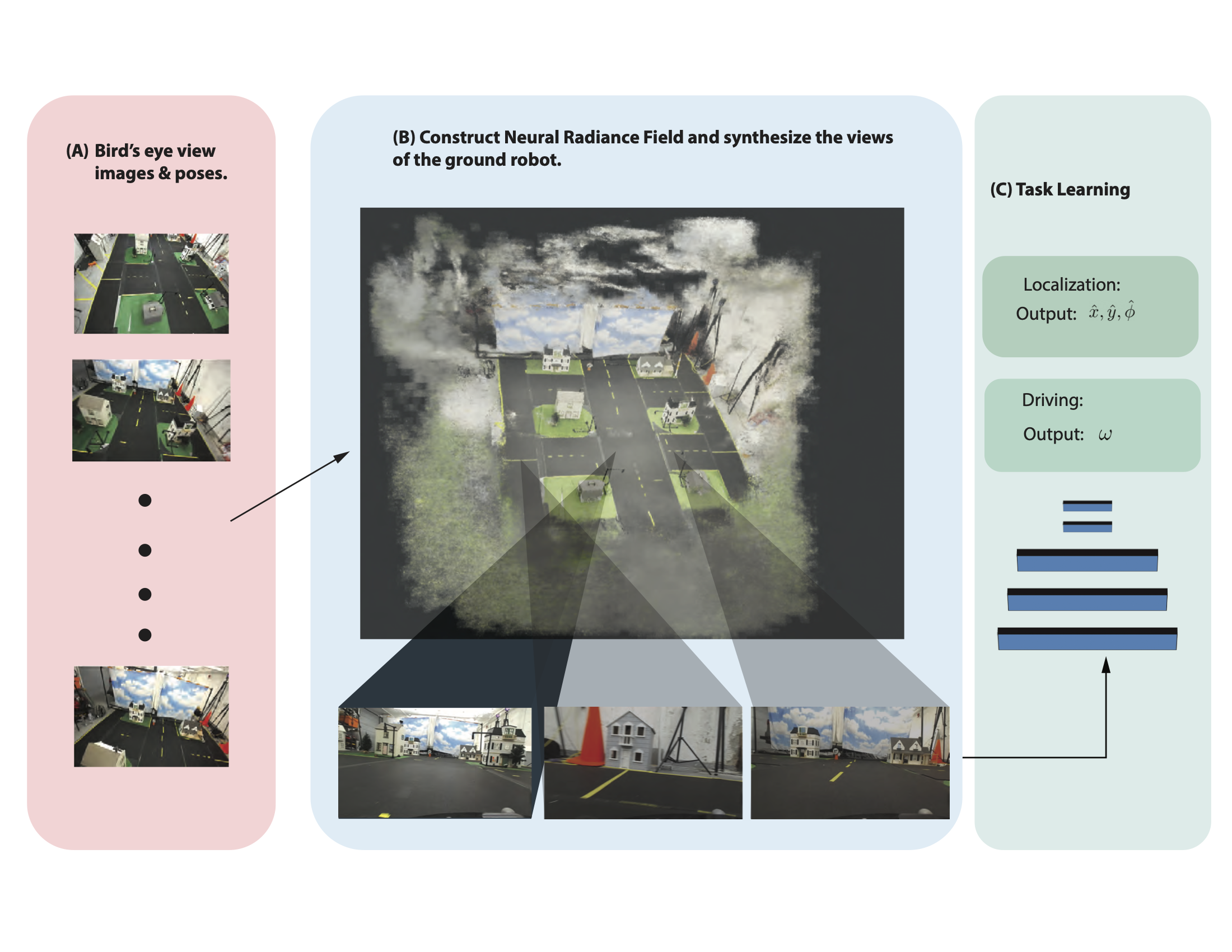}
\caption{An overview of our proposed method. (A) We assume that are we are given images and their corresponding poses from the bird's eye view. (B) We leverage a neural render field to compactly represent the density and color of the desired scene. The NeRF can be queried for poses along the road network. (C) We then learn two task relevant to autonomous driving: (i) visual localization and (ii) end-to-end driving.
}
\label{fig:pipeline}
\end{figure*}

\subsection{Problem Setup}
Let's denote images $I$ with the superscript $A$ representing a bird's eye view and the superscript $G$ represent a ground view.
The transform $_w^rT$ represents the transform from the robot to the world reference frame and $\mathcal{K}$ represent the camera matrix.
Given a sequence of images rendered from the point of view of the aerial robot represented by $I^A_{1, \cdots, n}$ and their locations $_w^rT_{1,\dots, n}$ we wish to find a policy $\phi$ that maps an image taken from the point of view of the ground vehicle $I^G$ to a control command comprising the steering $\omega$ and velocity $v$ command.
This requires the implicit construction of the scene represented by $c,\sigma$ where $c$ represents a function mapping the color to a physical location in the world and $\sigma$ represents a function mapping the probability of the location being occupied.
Figure \ref{fig:pipeline} shows the full pipeline of our proposed approach.

\subsection{Reconstruction}
\textbf{Model.} Given the sequence of images $I^A_{1,\cdots,n}$, we learn a neural radiance field using a multi-resolution hash approach \cite{muller2022instant}.
\newtext{As a backbone network for the NeRF, we use the method presented by \cite{muller2022instant} to learn two concatenated multilayer perceptron.}
The first learns the density i.e. a mapping $\sigma = \Psi(x;\theta)$ from the encoded position and direction of a projected ray to the density at that location.
The second learns the color and allowing for view-dependent color variation.
\newtext{We use an exponential moving average optimizer with a learning rate of $1e^{-3}$ and $\beta_1=0.9$ and $\beta_2=0.99$ for 5000 steps.}

\textbf{Adding road priors.} Since our method is reliant on cross view transfers and several occlusions exist in the aerial views, we enforce an additional prior on the depth and color $c_r$ of the road networks. 
We assume here that the location of the road network is available to us apriori.
We randomly include rays that are normal to the road surface with the predefined color to the training batch.
For each ray that is normal to the road surface, we first sample a position along that ray and include the position, view direction and pre-definied color into the training buffer.
We assume that the distance between the sampled ray position and the robot is equal to the height above ground of the aerial vehicle.

\textbf{Losses.} For each image in the training set we compute the loss of between the reconstructed image by projecting rays from the image location $T_w^I$ and the training images using a photometric error metric.
The road prior rays also contribute to the loss by computing the error between the predicted color and the desired road color.
Additionally, since the poses of the cameras are noisy we optimize over the poses of the cameras by computing the photometric error of projecting a nearby view onto the desired view.

\textbf{Algorithm.} In summary, the algorithm used for reconstruction from the bird's eye view image sequence is is presented in Algorithm~\ref{alg:reconstruction}. 
The input to the algorithm is the sequence of images and this stage, we produce a trained neural network that can output density and color given a ray.
The function ray takes a pixel and pose and returns the ray between them.
\begin{algorithm}
\caption{Procedure for reconstruction.}
\label{alg:reconstruction}
\begin{algorithmic}
\Require $^r_wT_{1,\cdots,n}, I^A_{1,\cdots,n}, N$
\For{$x = 1 \rightarrow N$}
\For{$k = 1 \rightarrow n$} 
\For{pixel in $I^G_k$} 
    \State $x,\theta, c$ = ray($^r_wT_k, \textrm{pixel})$ 
    \State Buffer $\gets (x, \theta, c)$ 
\EndFor
\For{pixel in road network}
    \State $T \gets $ random sample $\perp$ to road
    \State $x,\theta, c_r$ = ray($T, \textrm{pixel})$ 
    \State Buffer $\gets (x, \theta, c_r)$ \EndFor
\State Shuffle(Buffer)
\State TrainNerf(Buffer)
\EndFor
\EndFor
\end{algorithmic}
\end{algorithm}

\subsection{Data generation and models}
\textbf{Image Generation.}
Given the reconstruction, we can synthesize images at novel locations given a query pose.
We project rays consistent with the camera matrix $\mathcal{K}$ from each query location to get a set of RGB values and densities along the ray. 
Assuming the density is a log probability of the ray producing a ``hit", we can then sum the product of the predicted colors and the densities to get a color value for that pixel.
To generate adequate training data, we sample poses along the known road network. 
We take care to sample poses that would also violate driving rules (\eg~driving on the wrong side of the road) to provide data augmentation.
This is an advantage of using an intermediate reconstruction or ``simulation" since we can generate views that would not normally be present in collected driving data.
Prior work has already demonstrated the importance of this \cite{amini2020learning}.

\textbf{Control generation.}
In this work, we use the idea of pure-pursuit control as formulated in \cite{Coulter1992ImplementationOT}. 
However, our method is generic and can admit other types of control schemes in the policy training if desired. 
In the pure-pursuit algorithm a target path is known, a point on this path some fixed distance ahead of the vehicle is targeted, and a heading command is derived from the car's position and dimensions such that it will intersect this point within an acceptable margin. 
For this control scheme, we assume that the output velocity is constant. 
\newtext{An external rule based controller is used to change the velocities around obstacles.}
We also assume that the controller acts as a low pass filter on the steering commands to prevent over and under steering due to noisy inputs.

\begin{algorithm}
\caption{Procedure for task learning.}
\label{alg:policy-learning-procedure}
\begin{algorithmic}
\Require $^r_wT_{1,\cdots,n}, N, task$
\For{$x = 1 \rightarrow N$}
\For{$k = 1 \rightarrow n-1$} 
    \State $^r_wT_k \gets ^r_wT_k + perturb$ 
\For{pixel in $I^G_k$} 
    \State $x,\phi$ = projectRay($^r_wT_k)$ 
    \State pixel $\gets c(x,\phi)$
\EndFor
\If{task = visuo-motor}
\State $\omega, v \gets $ computeControl($^r_wT_k, ^r_wT_{k+1}$)
\State Buffer $\gets I^G_k, (\omega,v)$)
\Else
\State Buffer $\gets I^G_k, (^w_rT)$)
\EndIf
\EndFor
\State Shuffle(Buffer)
\State TrainModel(Buffer)
\EndFor
\end{algorithmic}
\end{algorithm}

\textbf{Model.} For both tasks under consideration, we use a five layer convolutional with 8 filters neural network with batch normalization on each layer and a rectified linear unit for activation to encode the images.
The output of the convolutional neural network is fed through two fully connected layer with 64 hidden units each to output the steering command for the driving task and to output the position and yaw for the localization task.
We implement the neural network in the pytorch framework \cite{NEURIPS2019_9015} with the ADAM optimizer \cite{adam-optimizer} with a learning rate of $1e^{-2}$ for 100 epochs.
\newtext{We stress that we intend here to keep the model size small to be feasibly deployed on the embedded platform on our robots but emperically we observe that these networks elicit the desired performance.}

\subsection{Tasks }
To demonstrate the cross-view transfer and it's applicability to autonomous driving, we choose the tasks of visual pose estimation and visuo-motor policy learning but stress that the method could be applicable to other tasks.
\newtext{We use the models described in the previous subsection.
The models are trained independently per-task and per-environment and do not share any parameters.
We also highlight that the second task is end-to-end and implicitly reasons about localization.}


\textbf{Visual pose estimation. } The first task we consider is that of estimating the robot pose given a single image $I^G$ and the relative transformation between the camera and the robot $^c_rT$.
For this task, we assume that the configuration of the environment is static and does not change between when the bird's eye view imagery was collected and the policy was deployed onboard the ground vehicle.
In this case, we train the neural network on all feasible locations of the ground robot for 100 epochs using Algorithm~\ref{alg:policy-learning-procedure}.
To train this model, we compute the loss using an L2 loss between the ground truth image pose and the predicted pose.
\newtext{Concretely, the goal here is to learn directly the desired steering angle of the car given the current camera image.}

\textbf{Visuo-Motor Policy Learning.} The next task under consideration is that of learning steering commands given the onboard imagery of the robot.
We assume that we are given a predefined trajectory that we would like to follow in the world frame.
Given the trajectory in the world frame, we sample a sequence of positions and orientations along that path and generate images from the reconstruction from the point of view of the car at those locations.
The procedure to train the model for 100 epochs is given in Algorithm~\ref{alg:policy-learning-procedure}.
To train this model, we compute the loss using an L1 loss between the desired steering and the predicted steering command.
\newtext{The goal of this model is to learn directly the desired steering angle of the car given the current camera image.}

%% file: experiments.tex
\section{Experiments}
\label{sec:experiments}

In this section, we detail the experimental setup and experiments to validate our method.
First, we discuss the experimental platform used for validation, the data collection pipeline and then the experiments for the two tasks: (i) localization and (ii) end-to-end driving.
We assume in this section, that a neural network is trained for each task and each configuration of the environment.

\subsection{Mini-city experimental platform}
\begin{figure}
    \centering
    \includegraphics[width=0.48\textwidth]{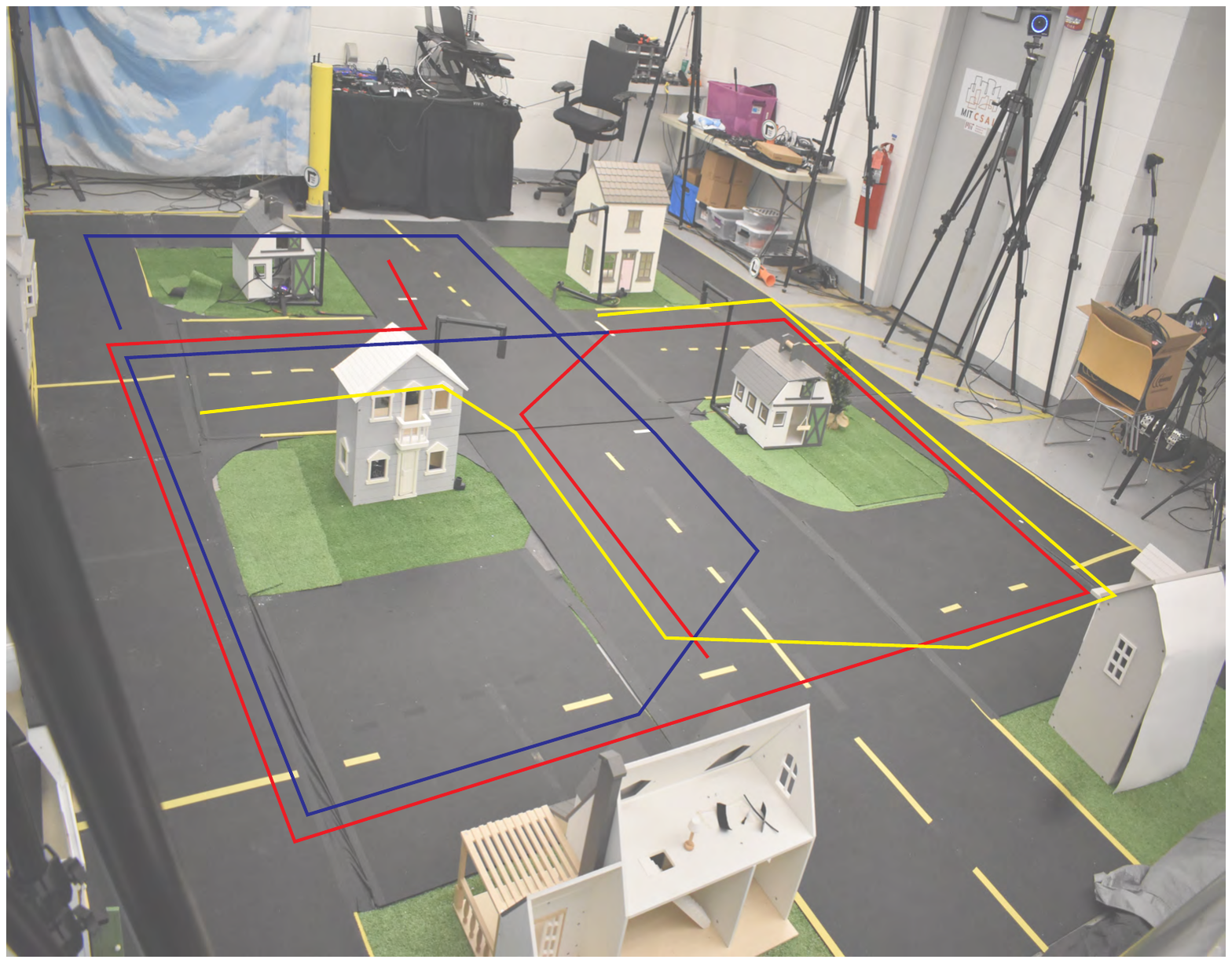}
    \caption{The figure shows the desired trajectories in the minicity environment.
    The first trajectory is shown in red, the second in blue, and the third in yellow.}
    \label{fig:trajectories}
\end{figure}

We use the mini-city experimental platform described in \cite{buckman2022evaluating} and visualized in Figure \ref{fig:trajectories}.
We deploy the learned policies onboard a custom built racecar built on the MIT racecar platform \cite{racecar}. 
The racecar carries a Jetson Xavier NX which is used to run the policies and a ZED 2i camera to acquire images and inertial measurements.
We use a Parrot Bebop 2 drone to collect images in the mini-city.
We employ 12 Optitrack Prime 41W cameras to estimate the poses of robots inside the minicity for associating the poses from the bird's eye view and for ground truth for the driving tasks.
\newtext{For each experiment, we collect approximately 100 images from the aerial view to generate the NeRF.}

\begin{figure} [!tbh]
    \centering
    \includegraphics[width=0.48\textwidth]{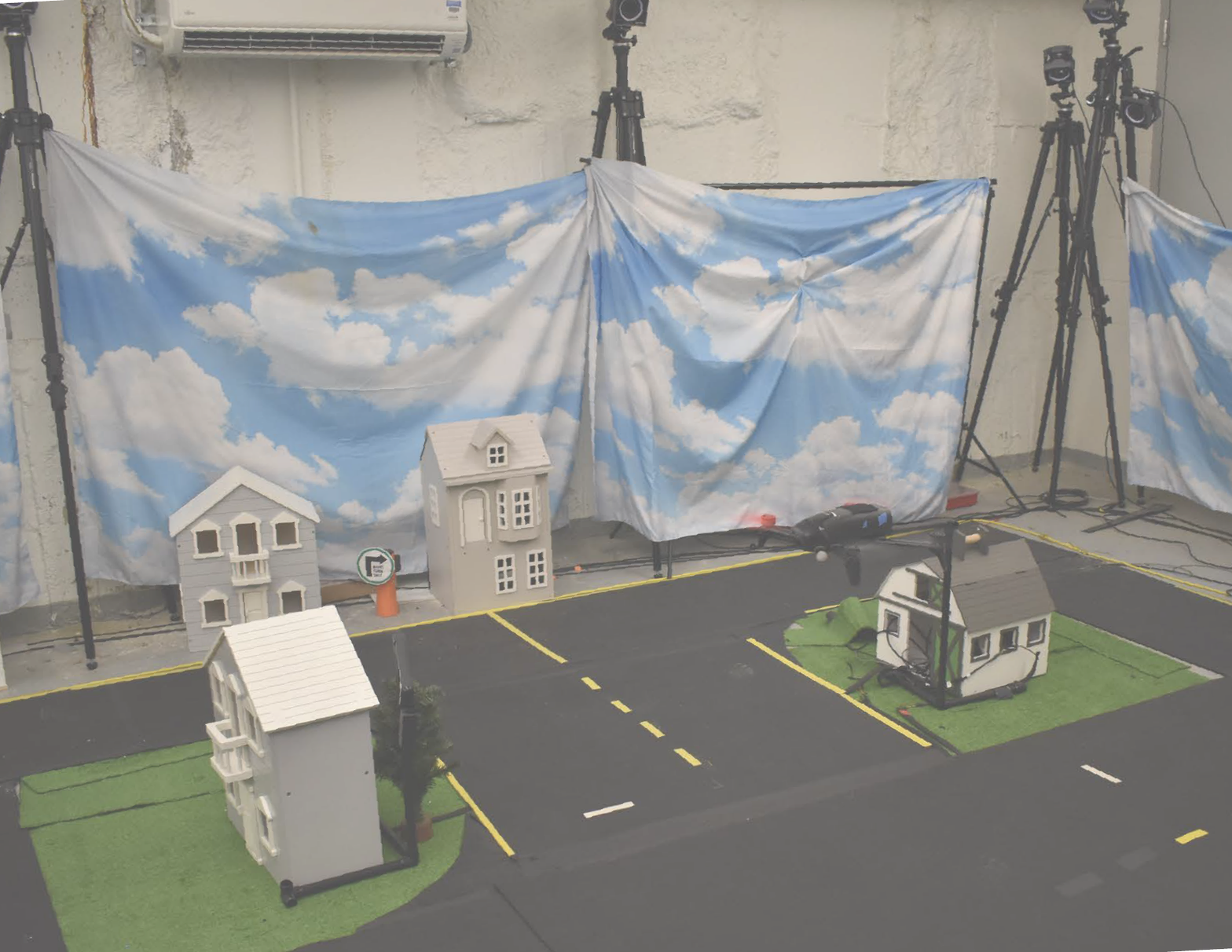}
    \caption{Experimental setup used. The figure shows an example configuration of the houses, the road network, the motion capture system and the Parrot Bebop drone used to collect data.}
    \label{fig:minicity}
\end{figure}

\subsection{Qualitative comparisons}
\begin{figure}
    \centering
    \begin{minipage}{.4\textwidth}
    \includegraphics[width=0.95\linewidth]{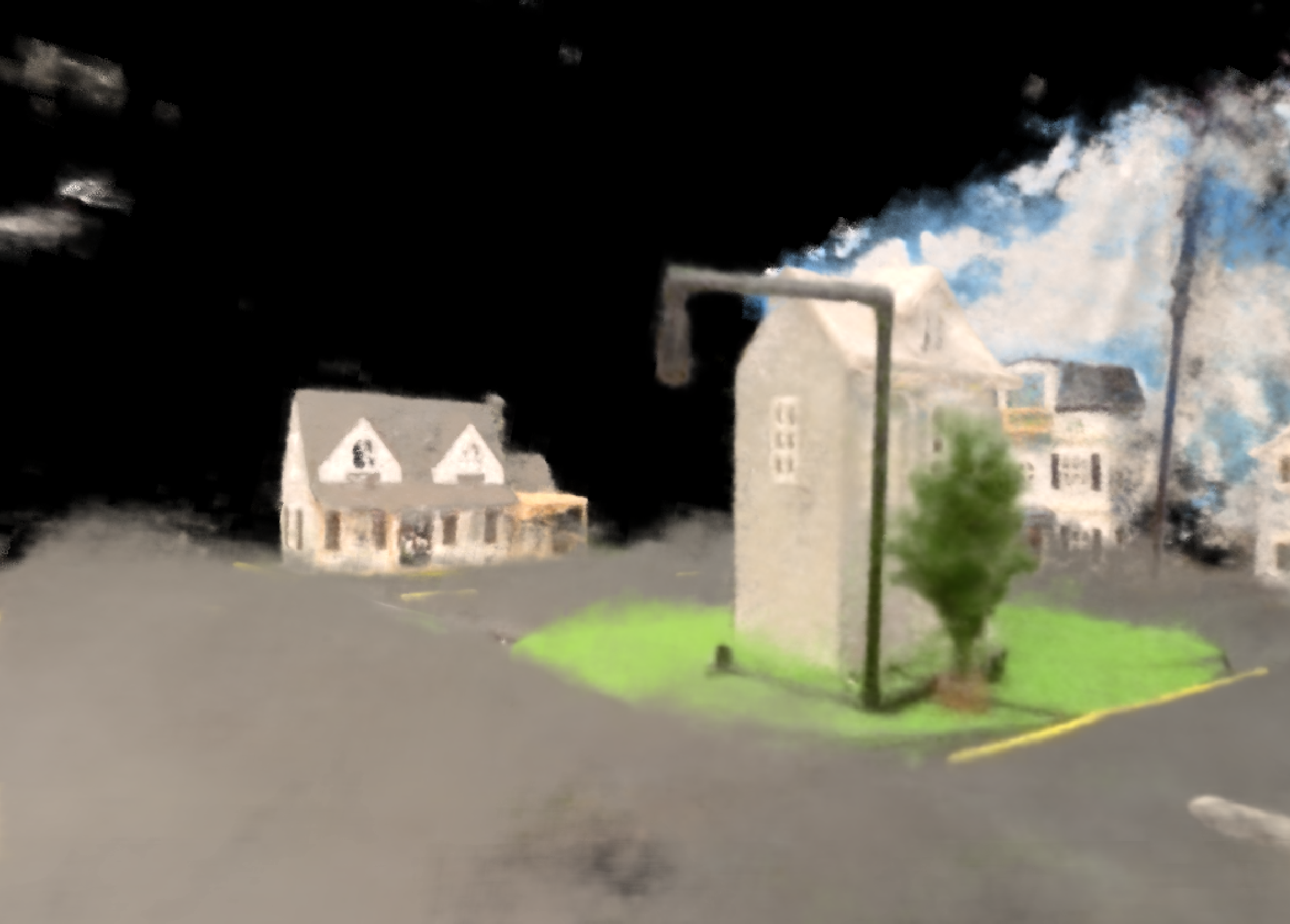}
    \end{minipage}
    ~
    \begin{minipage}{.4\textwidth}
    \includegraphics[width=0.95\textwidth]{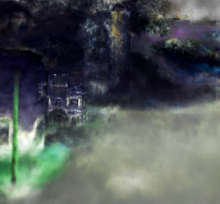}
    \end{minipage}
    \caption{Rendered image from the NeRF from the point of view of the ground robot with ground priors~(top) and without(bottom).}
    \label{fig:qualitative-nerf}
\end{figure}

First, we evaluate the visual quality of the cross view transfer using the NeRF as an intermediate representation.
We desire to study the effect of using the road priors on the intermediate reconstruction.
As can be seen in Figure~\ref{fig:qualitative-nerf}, we can see that the rendered images from the point of view of the ground robot captures details close to the ground and reasonable captures the geometry of environmental features such as houses.

\begin{figure*}[!tbh]
\includegraphics[trim={0 8cm 0 6cm},clip, width=\linewidth]{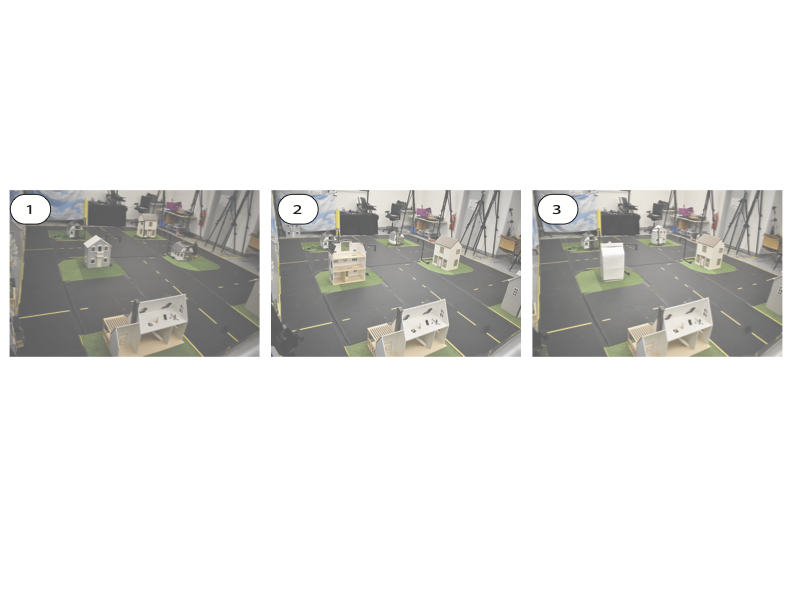}
\caption{The different configurations of the environment used for the visual localization task.}
\label{fig:scenarios}
\end{figure*}

\subsection{Visual Localization}
\textbf{Setup.} In this experiment, we measure the re-localization error using the localization policy. 
The ground robot is manually driven around the environment while running the localization network onboard. 
The imagery from the ZED camera on the ground robot is fed into the network and the output pose is recorded. 
We also recorded a filtered estimate that fuses the output pose from the neural network and the inertial measurements on the robot.
The different configurations of the environment used for the visual localization task is shown in Figure~\ref{fig:scenarios}.

\textbf{Metrics.} For this experiment, we show the root mean square error of estimating the poses along the trajectory for the position ($\|{x}-{x}_{ref}\|_2$) and orientation ($\|\psi-\psi_{ref}\|$) with respect to the ground truth collected using motion capture.

\textbf{Baseline.} As a baseline, we train our network on data collected from manually driving the ground robot around the environment.
The number of training images for both the baseline and our method is similar in the number of poses sampled along the trajectory.

\textbf{Results. } The results of the experiment are summarized in Table~\ref{tab:localization-results}.
As can be seen in the table, our method outperforms the baseline when the inertial measurements are added to the localization even though no ground view imagery is used while training. 
Our method is also comparable to the baseline when then inertial measurements are not included.

\begin{table}[!tbh]
\centering
\begin{tabular}{ c | c | c }

 Method & $\|{x}-{x}_{ref}\|_2$ [m] &  $\|\psi-\psi_{ref}\|$ [deg] \\ 
 \hline
 \multicolumn{3}{c}{\textbf{Scenario 1}} \\
\hline
 Baseline & 0.13 & 5.31 \\
 Ours & 0.17 & 7.82 \\  
 Ours + IMU & 0.11 & 2.26 \\
 \hline
\multicolumn{3}{c}{\textbf{Scenario 2}} \\
\hline
 Baseline & 0.15 & 7.89 \\
 Ours & 0.21 & 8.61 \\  
 Ours + IMU & 0.12 & 1.59 \\
 \hline
\multicolumn{3}{c}{\textbf{Scenario 3}} \\
\hline
 Baseline & 0.12 & 3.45 \\
 Ours & 0.18 & 7.44 \\  
 Ours + IMU & 0.09 & 1.12 \\

 \hline
\end{tabular}
\caption{The root mean square estimation error of the position and yaw angle of the ground robot with respect to the motion capture system.}
\label{tab:localization-results}
\end{table}

\subsection{End to End Control}
\textbf{Setup.} In this experiment, we assume that are we given a specific set of desired poses for the ground robot and we train the end-to-end driving network for that sequence of poses.
The desired trajectories are shown in Figure~\ref{fig:trajectories}.
We allow the network to specify the steering angle given the onboard imagery while keeping the nominal speed constant.
For this experiment, we record the pose of the driven trajectory and the ground truth from the motion capture system.

\textbf{Metrics.} We measure the number of interventions required while deploying the learnt policies and the root mean square error for pose ($\|\vert{x}-\vert{x}_{ref}\|_2$) and orientation ($\|\psi-\psi_{ref}\|$) from the pre-defined trajectory.

\textbf{Imitation.} The first baseline that we implement is pure imitation learning from collected datasets from the point of view of the ground robot.
We collect teleoperated data in the minicity and train the imitation policy using the collected dataset and test the performance by deploying the policy on the robot.
Since there is no domain shift, we expect this to perform reasonably with no need for additional sensors.

\textbf{Dataset + View Synthesis from ground view.} In the second baseline, we generate a NeRF from the collected datasets and add additional data by perturbing the poses of the robot in the dataset.
We include the predicted images from the NeRF for the perturbed path and the required control to bring the robot back to the nominal path in the training set \cite{bojarski2016end}.
For all methods, we use the same neural network with the only difference being the input data for fair comparison.
\newtext{This would yield a similar performance and setup as \cite{byravan2023nerf2real} since the domain of the training data would be similar to the task data.}

\textbf{Results. } The results of this experiment are summarized in Table~\ref{tab:end-to-end-driving}. 
As can be seen in the table, our method outperforms the baselines when no augmentation is used and outperforms the ground view augmentation when also augmenting with the bird's eye view.
The results also show that our method requires less interventions than the baseline.

\begin{table}[!tbh]
\centering

\begin{tabular}{ c | c | c | c}
Method & $\|{x}-{x}_{ref}\|_2$ [m] &  $\|\psi-\psi_{ref}\|$ [deg]  & Interventions \\
 \hline
 \multicolumn{4}{c}{\textbf{Trajectory 1}} \\
\hline
 Imit & 0.39 & 6.71 & 10 \\
 Imit + GV & 0.23 & 3.89 & 3 \\  
 Ours & 0.36 & 7.39 & 5 \\
 Ours + GV & \textbf{0.11} & \textbf{2.21} & \textbf{1} \\
 \hline
 \multicolumn{4}{c}{\textbf{Trajectory 2}} \\
\hline
 Imit & 0.27 & 13.8 & 5 \\
 Imit + GV & 0.14 & 7.29 &  1 \\  
 Ours & 0.20 & 7.47 & 2 \\
 Ours + GV & \textbf{0.09} & \textbf{3.73} & \textbf{0} \\
 \hline
\multicolumn{4}{c}{\textbf{Trajectory 3}} \\
\hline
 Imit & 0.34 & 14.16 &  8 \\
 Imit + GV & 0.24 & 10.60 &  2 \\  
 Ours & 0.16 & 10.75 & 3 \\
 Ours + GV & \textbf{0.12} & \textbf{3.57} & \textbf{1} \\
 \hline
\end{tabular}

\caption{The root mean square estimation error of the pose and the number of interventions during execution. Imit: Imitation from the user driven trajectories only. GV: Adding ground view augmentation through the NeRF.}
\label{tab:end-to-end-driving}
\vspace{-2em}
\end{table}